\documentclass[10pt,twocolumn,letterpaper]{article}

\usepackage[pagenumbers]{cvpr} 

\usepackage{graphicx}
\graphicspath{{Images/}}
\usepackage{amsmath}
\usepackage{amssymb}
\usepackage{booktabs}

\usepackage[pagebackref,breaklinks,colorlinks]{hyperref}

\usepackage[capitalize]{cleveref}
\crefname{section}{Sec.}{Secs.}
\Crefname{section}{Section}{Sections}
\Crefname{table}{Table}{Tables}
\crefname{table}{Tab.}{Tabs.}

\def\cvprPaperID{6836} 
\def\confName{CVPR}
\def\confYear{2022}

\begin{document}

\title{SL-CycleGAN: Blind Motion Deblurring in Cycles using Sparse Learning}

\author{Ali Syed Saqlain\\
NCEPU, Beijing, China\\
ncepu.edu.cn\\
{\tt\small 1164300014@ncepu.edu.cn}
\and
Li-Yun Wang\\
Portland State University, U.S.A\\
pdx.edu\\
{\tt\small liyuwang@pdx.edu}

\and
Fang Fang\\
NCEPU, Beijing, China\\
ncepu.edu.cn\\
{\tt\small ffang@ncepu.edu.cn}
}
\maketitle

\begin{abstract}
In this paper, we introduce an end-to-end generative adversarial network (GAN) based on sparse learning for single image blind motion deblurring, which we called SL-CycleGAN. For the first time in blind motion deblurring, we propose a sparse ResNet-block as a combination of sparse convolution layers and a trainable spatial pooler k-winner based on HTM (Hierarchical Temporal Memory) to replace non-linearity such as ReLU in the ResNet-block of SL-CycleGAN generators. Furthermore, unlike many state-of-the-art GAN-based motion deblurring methods that treat motion deblurring as a linear end-to-end process, we take our inspiration from the domain-to-domain translation ability of CycleGAN, and we show that image deblurring can be cycle-consistent while achieving the best qualitative results. Finally, we perform extensive experiments on popular image benchmarks both qualitatively and quantitatively and achieve the record-breaking PSNR of 38.087 dB on GoPro dataset, which is 5.377 dB better than the most recent deblurring method.    
\end{abstract}
\begin{figure}[ht]
    \centering
     \includegraphics[width=\columnwidth]{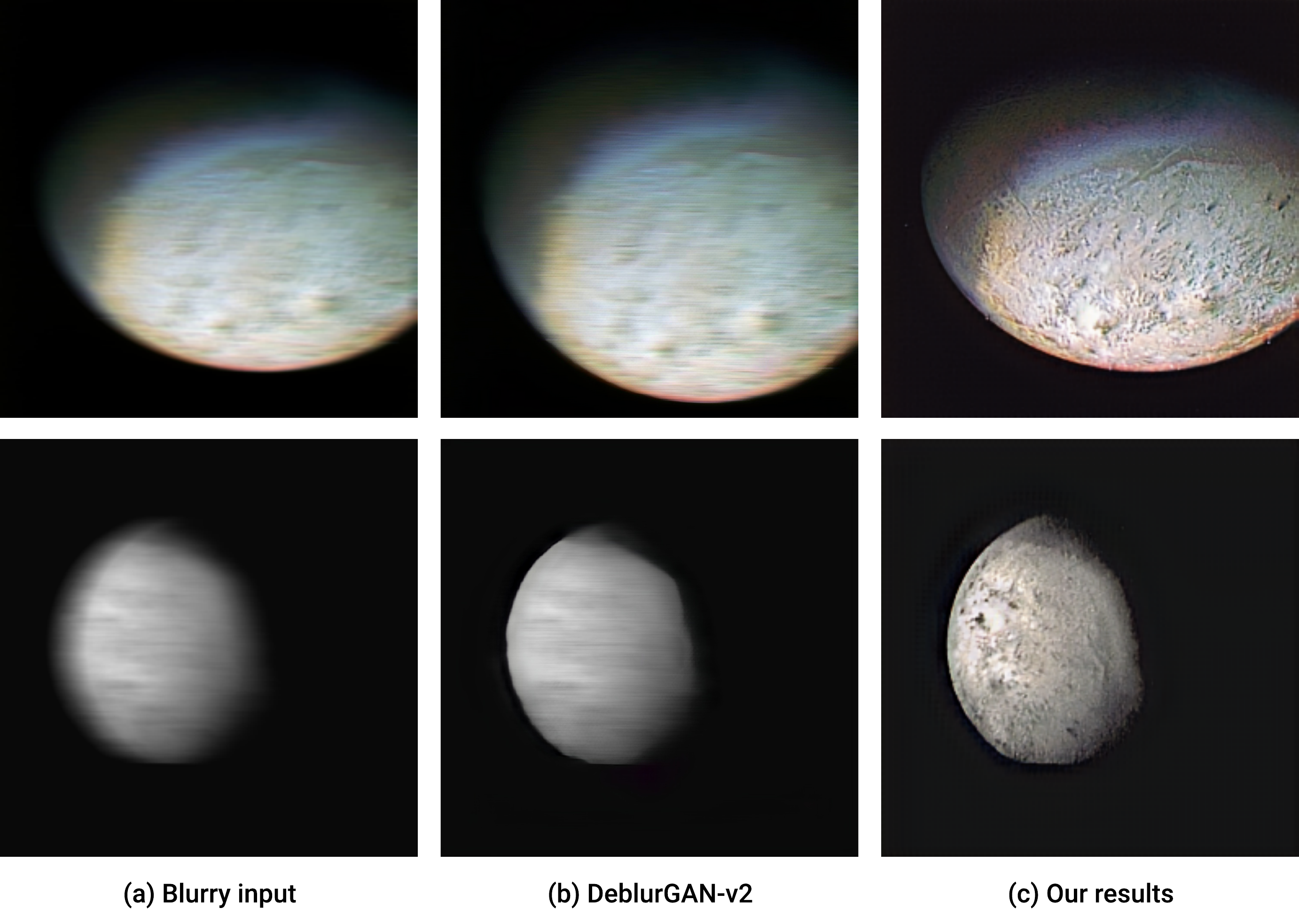}
\caption{Photos of Triton by Voyager, Neptune's satellite. Images taken from NASA's official website \cite{NASA}. From left to right: first column, the blurry input, second, the restored results from \cite{kupyn2019deblurgan} and finally, our restored results. The synthetic motion blur in the input images is of magnitude of 96 pixels with 0 degrees angle.}
  \label{fig1} 
\end{figure}
\section{Introduction}\label{intro}
Motion blur is one of the most recursive problems of paramount importance in the field of computer vision and digital photography. It is mainly caused by the streaking of fast moving objects in an image or video frames, also some of the other reasons could be the camera shake or long exposure time \cite{lagendijk2009basic, wang2014recent, yitzhaky1997identification, nayar2004motion, shan2008high}. An interesting way to understand motion blur is to understand the concept of relative motion, \eg, motion of an object relative to the observer in an instance of time. During a single exposure time, the image captured by the camera, especially when an object is moving in the captured image may represent a scene over a continuous interval of time. Such motion of an object in a captured image, causes motion blur artifacts or more specifically displacement of pixels. Removal of motion blur from the images, is to obtain the clean sharp images from the blurry inputs by minimizing the mismatch between the latent sharp image and the restored sharp image. Considering the fact, that motion blur in real-world is commonly shift-variant or non-uniform in nature, and the depth or the density of blur may fluctuate over different regions of an image \cite{bardsley2006blind, cho2007removing, hirsch2011fast, couzinie2013learning}. The earlier blind deconvolution methods assume the estimation of unknown blur kernel by studying image priors \cite{krishnan2009fast, levin2011efficient, zoran2011learning, xu2013unnatural, werlberger2010motion, zhang2015image, pan2016blind}, Weiner deconvolution \cite{wiener1950extrapolation, krishnan2009fast}, or Richardson-Lucy bayesian approach \cite{richardson1972bayesian}. However, such outdated methods inevitably need the handcrafted image priors, while not to mention the cost of computational resources and the exponential increase in complexity. \par
Thanks to the recent advancements in deep learning, and its tremendous feature learning ability from training to testing in blind image deblurring task \cite{schuler2015learning, sun2015learning, nah2017deep, tao2018scale, zhang2018dynamic}, we do not need to rely on conventional approaches anymore. Especially, considering the revolution that GANs \cite{goodfellow2014generative} have brought not only in general-purpose computer vision tasks \cite{karras2019style, karras2020analyzing}, but also in blind motion deblurring task \cite{kupyn2018deblurgan, kupyn2019deblurgan, shao2020deblurgan+, asim2020blind, zhang2021deep}. Despite the colossal success of GANs in blind motion deblurring, the quality of restored images from the blurry inputs is still straggling. No doubt, either scale-wise stacking of convolution layers \cite{sun2015learning, gong2017motion, noroozi2017motion, nah2017deep, schuler2015learning, tao2018scale, zhang2018dynamic, gao2019dynamic, stacked} or versions of GAN-based DeblurGAN models \cite{kupyn2018deblurgan, kupyn2019deblurgan} have significantly improved the performance of restoration both in terms of qualitative and quantitative analysis. \par 
To address the problem of non-uniform blind image deblurring, and to propose such a GAN-based blind image motion deblurring network, that, unlike other GAN-based models does not treat the restoration of sharp image as a linear end-to-end process from input to output. Instead, our proposed approach (SL-CycleGAN) treats blind motion deblurring as a domain-to-domain translation problem. What's even more interesting is, we take the inspiration for this research from the sparse representation learning of \cite{ahmad2019can}, not only this, but we also combine Hawkins \etal \cite{hawkins2011cortical} research on Hierarchical temporal memory (HTM) and ``A thousand brains: A new theory of intelligence'' by Jeff Hawkins \cite{hawkins2021thousand}. By observing other GAN-based state-of-the-art methods for blind motion deblurring \cite{kupyn2018deblurgan, kupyn2019deblurgan, cai2020dark}, the results achieved by our proposed framework outperform state-of-the-art motion deblurring methods, and speak for themselves both qualitatively and quantitatively. \cref{fig1} shows the supreme reconstruction ability of our proposed method against DeblurGAN-v2 \cite{kupyn2019deblurgan} on low-light space images. \par
Our contributions in this research paper are summarized as follow:
\begin{itemize}
    \item \textbf{The Framework:} There is no denial that, GAN-based models are notorious in nature when it comes to problems such as mode-collapse and vanishing gradient \cite{arjovsky2017towards, radford2015unsupervised}. Therefore, a thoughtful choice of network that is able to tackle such issues is of vital importance. We adopt CycleGAN \cite{zhu2017unpaired} for its amazing ability of domain-to-domain translation. Unlike, other GAN-based models for blind motion deblurring, our proposed network is cycle-consistent, that means, not only the generators of our network are able to deblur the blurry input but also they are able to reconstruct synthetic non-uniform motion blur similar to the original blurry input. 
    \item \textbf{Sparse Convolutions:} Our second contribution is the adoption of intrinsic advantages of high dimensional sparse representation through sparse convolutions, similar to \cite{ahmad2019can}. The primary reason why we choose sparse convolutions over standard convolutions layers in our proposed generator architecture is, sparse representations are more robust to noise and interference.   
    \item \textbf{HTM:} Hierarchical Temporal Memory (HTM) is an algorithm that models how neocortex of a human brain performs complex world calculations such as understanding of visual patterns, context of spoken language, perceiving the information through touch and other sensory organs \cite{hawkins2011cortical}. Our final contribution is, utilizing a trainable HTM spatial pooler such as k-winner \cite{ahmad2019can} to replace non-linearity ReLU$\left( \cdot  \right)$ with k-winner in the residual-block of the generator network. The reason of k-winner as a replacement for classic ReLU is, it is naturally more robust to variance in noise and interference from random signals. In addition, k-winner constraints the output of each layer to the most active non-zero units. 
\end{itemize}
\section{Related work}\label{related}
\subsection{Motion Deblurring}
Earlier methods treat blind motion image deblurring as an image deconvolution problem \cite{shan2008high,fergus2006removing, cho2009fast, xu2010two}. Similarly, sparse-based methods before the introduction of CNNs explore the sparse image gradients in the input blurry images \cite{krishnan2011blind, levin2011efficient, pan2016l_0, perrone2016logarithmic, sun2013edge, xu2010two, xu2013unnatural}. Meanwhile, other similar motion deblurring approaches focus more on the advantages of patch-wise estimation \cite{michaeli2014blind} and estimation of dark channel image priors \cite{pan2016blind}. However, these conventional image deconvolution methods assume the blur to be uniform, while real-world blur is mostly non-uniform or shift-variant. Since the introduction of deep learning and CNNs, the blind motion deblurring community has seen exponential improvement in the quality of restored sharp images such as \cite{sun2015learning, gong2017motion, noroozi2017motion}. Nah \etal. \cite{nah2017deep} proposed a deep scale-wise convolution network for dynamic scene motion deblurring. Unlike conventional deconvolution methods, \cite{nah2017deep} eliminates the need of knowing explicit blur kernel in advance. Similarly, Schuler \etal. \cite{schuler2015learning} proposed deep CNN network for blind motion deblurring in a coarse-to-fine scheme. Tao \etal. \cite{tao2018scale} proposed a encoder-decoder scale-wise recurrent network architecture for blind motion deblurring. Zhang \etal. \cite{zhang2018dynamic} proposed a sequential CNN architecture for dynamic scene deblurring, while achieving impressive results in comparison with \cite{nah2017deep} and \cite{tao2018scale}. Gao \etal. \cite{gao2019dynamic}   proposed a parameter selective sharing scheme, and a multi-scale encoder-decoder model with nested skip connections for dynamic scene deblurring. Cai \etal. \cite{cai2020dark} proposed a dynamic scene motion deblurring network that investigates the dark and bright channel image priors in the input blurry images. 
\begin{figure*}[t]
    \centering
    \includegraphics[width=\textwidth]{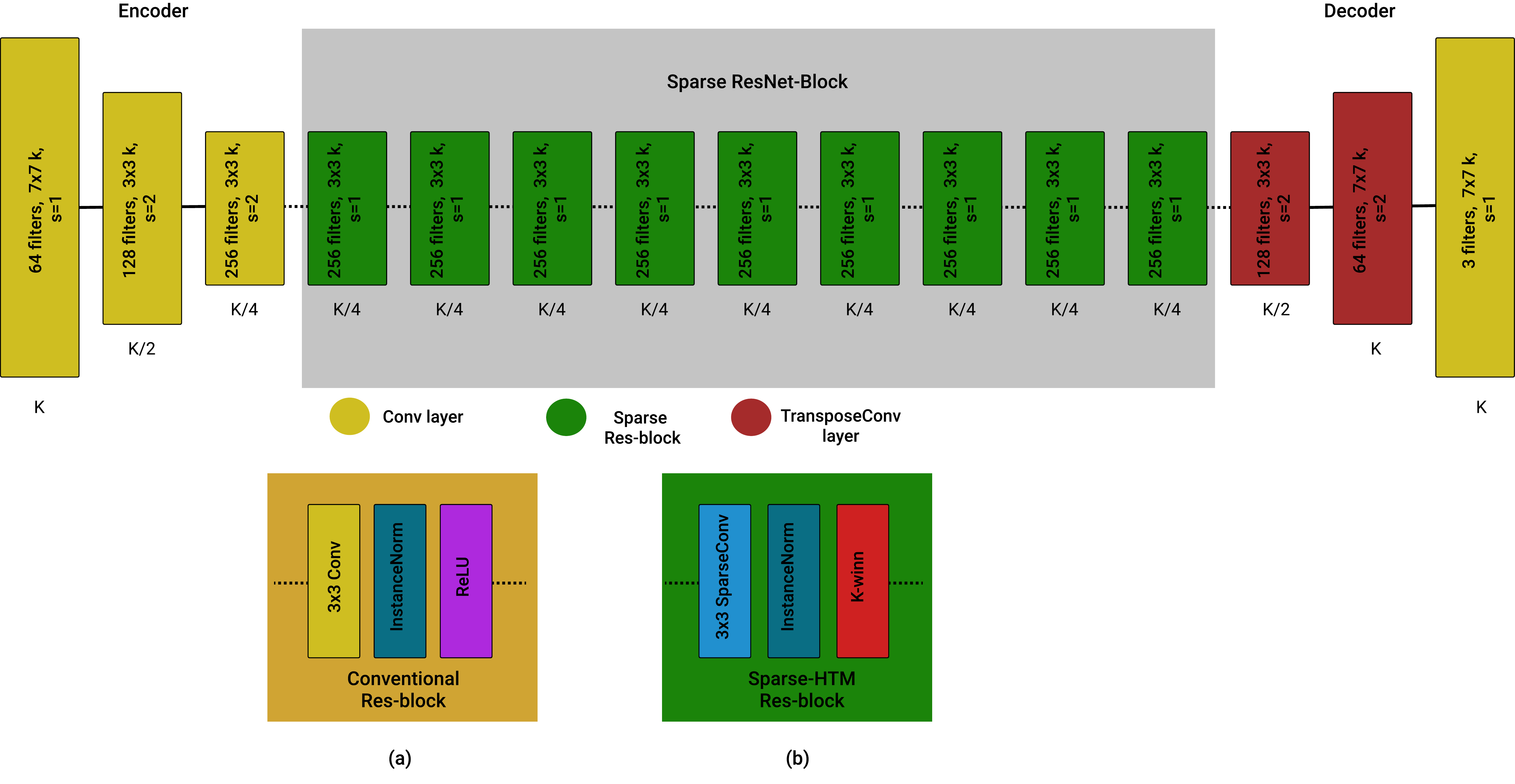}
    \caption{Architecture of SL-CycleGAN generators. The encoder block of the generator contains strided convolutions layers with stide of $\frac{1}{2}$. Each convolution layer in the encoder block and TransposeConv layer in the decoder block is followed by the InstanceNorm and non-linearity ReLU. Besides, the generators network contains 9-residual blocks. \cref{fig2}a shows the conventional Res-block architecture with Conv layers, InstanNorm, and ReLU, while \cref{fig2}b shows the modified Res-block with Sparse-Conv layer, InstanceNorm, and ReLU is replaced by HTM k-winn. Each Sparse Res-block contains a Sparse-conv layer, InstanceNorm, and k-winn.}
    \label{fig2}
\end{figure*}
\subsection{GANs for Motion Deblurring}
Generative adversarial networks (GANs) \cite{goodfellow2014generative} since their introduction, have been widely used for computer vision tasks. A GAN architecture is a deep neural network that consists of mainly a generator $G$ and a discriminator $D$. A generator network that generates a fake generated sample $G(z)$, normally takes a random noise vector $z$ as an input, more specifically in a deblurring scenario, an input blurry image is passed through the generator network and it outputs a fake sharp version of the input image. While the discriminator network acts as a  classifier by discriminating between the real data sample $x$ and the generated sample $G(z)$. Both of these adversaries play a minimax game and keep getting better and better. Theoretically, the main goal of such adversarial network is to approximate the generated distribution ${{p}_{z}}$ to the real data distribution ${{p}_{d}}$. The minimax objective function for GANs can be formulated as, 
\begin{equation}\label{eq1}
    \underset{G}{\mathop{\min }}\,\underset{D}{\mathop{\max }}\,\left[ {{\mathbb{E}}_{x\sim{{p}_{d}}}}\log (D(x)) \right]+\left[ {{\mathbb{E}}_{z\sim{{p}_{z}}}}\log (1-D(G(z))) \right]
\end{equation}\par
Based on the success of GANs, Kupyn \etal. \cite{kupyn2018deblurgan} proposed a conditional GAN-framework (DeblurGAN) for single image blind motion deblurring. DeblurGAN consists of a generator and a discriminator network for deblurring task, while utilizing Wasserstein loss function \cite{arjovsky2017wasserstein} and optimization criteria with an additional gradient penalty to tackle GAN related isssues \cite{gulrajani2017improved}. Kupyn \etal. \cite{kupyn2019deblurgan} proposed the improved version of previous DeblurGAN \cite{kupyn2018deblurgan}, called DeblurGAN-v2. The DeblurGAN-v2 modified the original architecture of the generator network of DeblurGAN by incorporating the Feature pyramid network (FPN) \cite{lin2017feature} for improved quality. Similarly a relativistic local and global discriminator network is introduced \cite{jolicoeur2018relativistic} with Inception-ResNet-v2 \cite{szegedy2017inception} as a backbone of the network. Shao \etal. \cite{shao2020deblurgan+} proposed a GAN-based deblurring framework that explores the dark and bright channel image priors. Asim \etal. \cite{asim2020blind} proposed a blind image deblurring network based on deep generative priors. However, their proposed method lack the experimental analysis on real-world benchmarks for deblurring. Zhang \etal. \cite{zhang2021deep} proposed a image deblurring and denoising network by combining the noisy and blurry image pairs acquired in a burst. Similarly, Lin \etal. \cite{lin2019tell} deployed a GAN-based framework for blind dynamic scene deblurring. Several other approaches either exploit the network architecture in scale-wise convolutions \cite{aljadaany2019douglas, zhang2019gated, jiao2017formresnet}, or the deep generative and discriminative priors \cite{ren2020neural, li2018learning} for dynamic scene blind motion deblurring. 
\section{SL-CycleGAN Network Architecture}
The detailed architecture of SL-CycleGAN generators is demonstrated in \cref{fig2}. Given a pair of blurry and sharp images $\left\{ {{x}_{i}} \right\}_{i=1}^{N}\in {{X}_{blurry}}$ and $\left\{ {{y}_{i}} \right\}_{j=1}^{M}\in {{Y}_{sharp}}$, the generators of SL-CycleGAN learn the translations from ${{X}_{blurry}}$ to ${{Y}_{sharp}}$. Taking the inspiration from the original CycleGAN model \cite{zhu2017unpaired}, SL-CycleGAN also introduces two generator networks ${{G}_{X}}$ and ${{G}_{Y}}$. Whereas, ${{G}_{X}}$ learns the translation function from ${X}\to {Y}$ such that ${{G}_{X}}:{{X}_{blur}}\to {{Y}_{sharp}}$. Similarly, the second generator ${{G}_{Y}}$ learns the mapping from ${Y}\to {X}$ such that ${{G}_{Y}}:{{Y}_{sharp}}\to {{X}_{blur}}$. A pair of adversarial discriminators ${{D}_{X}}$ and ${{D}_{Y}}$ are also proposed. While ${{D}_{X}}$ learns to differentiate between the blurry input ${{x}_{i}}$ and the translated image ${{G}_{Y}}(\hat{y})$. Similarly, ${{D}_{Y}}$ differentiates between the latent sharp image ${{y}_{i}}$ and the translated image ${{G}_{X}}(\hat{x})$. The architecture of our discriminator networks is similar to the PatchGAN $70\times 70$ discriminator \cite{isola2017image}.   
\subsection{Cycle-consistent Deblurring}
As mentioned earlier in \cref{intro}, GANs have been known for mode collapse. The main reason behind mode collapse in GANs, is its adversarial nature and the choice of the objective function for optimization purposes. Theoretically speaking, a generator function that maps ${{G}_{X}}:{{X}_{blur}}\to {{Y}_{sharp}}$ outputs a distribution of translated image ${{p}_{data}}(\hat{y})$ such that the output image $\hat{y}$ is similar to the original sharp image $y$. The assumption that the generated probability distribution ${{p}_{data}}(\hat{y})$ strictly correlates to the original data distribution ${{p}_{data}}({y})$ requires the generator ${G}_{X}$ to be stochastic in nature \cite{goodfellow2014generative}. However during inference, such theoretical assumption does not assure that the generator will learn meaningful translations without being the victim of mode collapse. \par
In order to avoid mode collapse in generators and to improve the optimization ability of the network, Zhu \etal. \cite{zhu2017unpaired} argued that the adversarial objective function of generators should be coupled with the term ``cycle-consistent''. The cycle-consistency term ensures that the generators ${{G}_{X}}$ and ${{G}_{Y}}$ are the inverse mapping functions of each other. It can be defined as, 
\begin{equation}\label{eq2}
    \begin{split}
        {{L}_{cycle}}\text{(}{{G}_{X}},{{G}_{Y}})={{\mathbb{E}}_{x\sim {{p}_{data}}(x)}}\left[ {{\left\| {{G}_{Y}}({{G}_{X}}(\hat{x}))-x \right\|}_{1}} \right]\\+{{\mathbb{E}}_{y\sim {{p}_{data}}(y)}}\left[ {{\left\| {{G}_{X}}({{G}_{Y}}(\hat{y}))-y \right\|}_{1}} \right]
    \end{split}
\end{equation}
where ${{L}_{cycle}}$ in \cref{eq2} represents the L1 norm, ${{p}_{data}(x)}$ and ${{p}_{data}(y)}$ represent the distributions of blurry and sharp images. \par
Unlike our close GAN-based competitors for blind motion deblurring \cite{kupyn2018deblurgan, kupyn2019deblurgan, shao2020deblurgan+, zhang2021deep}, we consider cycle-consistency an essential factor for blind motion deblurring, and show that our network outperforms all the state-of-the-art methods in blind motion dynamic scence deblurring in \cref{exp}.  
\subsection{Sparse Convolutions and HTM}
The history behind sparse representations is nothing new, in fact, Olshausen \etal. \cite{olshausen1997sparse} showed that deploying sparse embeddings and sparse objective functions in encoders can lead to the representations that are similar to the learnt representations in primate visual cortex. Similarly, Chen \etal. \cite{chen2018sparse} developed hierarchical sparse representations that are similar to hierarchical feature detectors. The weights for each unit in sparse convolution layers in our Resnet architecture are randomly sampled from a sparse subset of the previous layer. Additionally, the output of each layer is bounded to only $k$ most non-zero active units. The number of non-zero products in each layer is (sparsity of layer $l$)$\times$(sparse weights of layer $l + 1$). \cref{fig2}a represents a conventional arrangement of a residual-block, where each conv layer is followed by InstanceNorm and ReLU as an activation function. \cref{fig2}b is our modified structure for residual-block, which we called Sparse ResNet-block. To select the most active $k$ non-zero units and for each unit to be equally active in order to be robust to noise and interference, boosting techniques are applied to sparse convolution layers which can be defined as, 
\begin{equation}\label{eq3}
    c_{i}^{l}(t)=(1-\alpha )c_{i}^{l}(t-1)+\alpha \cdot \left[ i\in topIndice{{s}^{l}} \right]
\end{equation}
\cref{eq3} represents the HTM boosting duty cycle through k-winner, which calculates the running average of each active unit cycle. Where $c_{i}^{l}(t)$ is the unit duty cycles for each unit $i$ in layer $l$ at time $t$. The boosting coefficient $b_{i}^{l}={{e}^{\beta ({{{\hat{a}}}^{l}}-c_{i}^{l}(t))}}$ is then measured for each unit based on the target and current average duty cycle. Where ${{{\hat{a}}}^{l}}$ denotes the number of units that are expected to be active, while the boosting factor $\beta$ is a positive parameter responsible for controlling the strength of boosting. \par
In order to construct sparse convolutions with HTM k-winner in Resnet architecture, k-winner is applied to the output of InstanceNorm in each residual-block respectively with stride of 1 and kernel size of $3\times3$. 

\subsection{Loss functions}
In this section, we discuss the loss functions for our proposed SL-CycleGAN, the overall loss function is the combination of three different loss functions. \par
\textbf{Adversarial loss:} The adversarial objective functions is an essential component for blind motion deblurring in GANs. The classic Jensen-Shannon divergence (JSD) based  minimax loss function for GANs is proposed by \cite{goodfellow2014generative} is defined as in \cref{eq1}. However, the objective fucntion in \cref{eq1} suffers from the serious issues such as mode collapse and vanishing gradient. Thus, a conventional minimax objective function is not a good choice for our blind motion deblurring task. Instead, we choose the objective function of \cite{gulrajani2017improved} with gradient penalty term. The adversarial functions of our proposed network can be defined as follow, 
\begin{equation}\label{eq4}
    \begin{split}
        {{L}_{adv}}({{G}_{X}},{{D}_{Y}},{{X}_{blur}},{{Y}_{sharp}})={{\mathbb{E}}_{y\sim {{p}_{data}}(y)}}\left[ {{D}_{Y}}(y) \right]\\-{{\mathbb{E}}_{x\sim {{p}_{data}}(x)}}\left[ {{D}_{Y}}({{G}_{X}}(x)) \right],
    \end{split}
\end{equation}
\begin{equation}\label{eq5}
    \begin{split}
        {{L}_{adv}}({{G}_{Y}},{{D}_{X}},{{Y}_{sharp}},{{X}_{blur}})={{\mathbb{E}}_{x\sim {{p}_{data}}(x)}}\left[ {{D}_{X}}(x) \right]\\-{{\mathbb{E}}_{y\sim {{p}_{data}}(y)}}\left[ {{D}_{X}}({{G}_{Y}}(y)) \right]
    \end{split}
\end{equation}
where ${{G}_{X}}$ and ${{G}_{Y}}$ are the inverse mapping functions of each other. The adversarial functions for both the generators and discriminators in \cref{eq4} and \cref{eq5} are combined along with cycle-consistency loss ${{L}_{cycle}}$ from \cref{eq3} during the inference. \par
\textbf{Perceptual loss:} We observe that by incorporating only adversarial and cycle-consistency loss, the quality of the restored images is slightly degraded. To further improve the quality of restored images, we adopt the perceptual loss of pre-trained VGG-19 by Jonhnson \etal. \cite{johnson2016perceptual}. The perceptual loss can be defined as, 
\begin{equation}\label{eq6}
    {{L}_{perc}}=\frac{1}{{{W}_{i,j}}{{H}_{i,j}}}\sum\limits_{w=1}^{{{W}_{i,j}}}{\sum\limits_{h=1}^{{{H}_{i,j}}}{{{\left( {{\phi }_{i,j}}{{({{I}_{S}})}_{w,h}}-{{\phi }_{i,j}}{{({{G}_{\theta }}({{I}_{B}}))}_{w,h}} \right)}^{2}}}}
\end{equation}
where ${{H}_{i,j}}$ and ${{W}_{i,j}}$ in \cref{eq6} indicate the height and width of the conv3$\times $3 layers in the pre-trained VGG19 network. ${{\phi }_{i,j}}$ indicates the obtained feature maps by the j-th convolution layer after the activation function and before the i-th maxpooling layer. ${{I}_{S}}$ and ${{G}_{\theta }}({{I}_{B})}$ represent the real sharp and the restored deblurred images.\par
\textbf{Overall Loss Function:}
The overall loss function for proposed SL-CycleGAN can be defined as, 
\begin{equation}\label{eq7}
    {{L}_{SL-CycleGAN}}={{L}_{adv}}+{{\lambda}_{cyc}}{{L}_{cycle}}+{{\lambda }_{perc}}{{L}_{perc}}
\end{equation}
where ${{\lambda}_{cyc}}$ represents the relative coefficient of adversarial functions for ${G}_{X}$ and ${G}_{Y}$. ${\lambda }_{perc}$ is the hyper-parameter for perceptual loss ${{{L}}_{perc}}$. 
\section{Experimental evaluation}\label{exp}
\begin{table}[ht]
\centering
\resizebox{\columnwidth}{!}{%
\begin{tabular}{@{}clllclc@{}}
\toprule
Method                                &  & Year                     &  & PSNR                 &  & SSIM                 \\ \midrule
DeepDeblur \cite{nah2017deep}                            &  & \multicolumn{1}{c}{2016} &  & 30.12                &  & 0.9021               \\
\multicolumn{1}{l}{}                  &  &                          &  & \multicolumn{1}{l}{} &  & \multicolumn{1}{l}{} \\
DeblurGAN \cite{kupyn2018deblurgan}                          &  & \multicolumn{1}{c}{2018} &  & 28.70                &  & 0.958                \\
\multicolumn{1}{l}{}                  &  &                          &  & \multicolumn{1}{l}{} &  & \multicolumn{1}{l}{} \\
DeblurGAN-v2-Inception \cite{kupyn2019deblurgan}               &  & \multicolumn{1}{c}{2019} &  & 29.55                &  & 0.934                \\
\multicolumn{1}{l}{}                  &  &                          &  & \multicolumn{1}{l}{} &  & \multicolumn{1}{l}{} \\
DeblurGAN+ \cite{shao2020deblurgan+}                           &  & \multicolumn{1}{c}{2020} &  & 28.62                &  & 0.959                \\
\multicolumn{1}{l}{}                  &  &                          &  & \multicolumn{1}{l}{} &  & \multicolumn{1}{l}{} \\
DBGAN \cite{zhang2020deblurring}                                &  & \multicolumn{1}{c}{2020} &  & 31.10                &  & 0.9424               \\
\multicolumn{1}{l}{}                  &  &                          &  & \multicolumn{1}{l}{} &  & \multicolumn{1}{l}{} \\
RNNDeblur \cite{zhang2018dynamic}                            &  & \multicolumn{1}{c}{2018} &  & 29.1872              &  & 0.9306               \\
\multicolumn{1}{l}{}                  &  &                          &  & \multicolumn{1}{l}{} &  & \multicolumn{1}{l}{} \\
SRN-Deblur \cite{tao2018scale}                           &  & \multicolumn{1}{c}{2018} &  & 30.26                &  & 0.9342               \\
\multicolumn{1}{l}{}                  &  &                          &  & \multicolumn{1}{l}{} &  & \multicolumn{1}{l}{} \\
DBCPeNet \cite{cai2020dark}                           &  & \multicolumn{1}{c}{2020} &  & 31.10                &  & 0.945                \\
\multicolumn{1}{l}{}                  &  &                          &  & \multicolumn{1}{l}{} &  & \multicolumn{1}{l}{} \\
MTRNN \cite{park2020multi}                                &  & \multicolumn{1}{c}{2019} &  & 31.15                &  & 0.945                \\
\multicolumn{1}{l}{}                  &  &                          &  & \multicolumn{1}{l}{} &  & \multicolumn{1}{l}{} \\
DMPHN \cite{stacked}                                &  & \multicolumn{1}{c}{2019} &  & 31.50                &  & 0.9483               \\
\multicolumn{1}{l}{}                  &  &                          &  & \multicolumn{1}{l}{} &  & \multicolumn{1}{l}{} \\
SRN+PSS+NSC \cite{gao2019dynamic}                          &  & \multicolumn{1}{c}{2019} &  & 31.58                &  & 0.9478               \\
\multicolumn{1}{l}{}                  &  &                          &  & \multicolumn{1}{l}{} &  & \multicolumn{1}{l}{} \\
Learning Even-Based Motion Deblurring \cite{jiang2020learning} &  & \multicolumn{1}{c}{2020} &  & 31.79                &  & 0.949                \\
\multicolumn{1}{l}{}                  &  &                          &  & \multicolumn{1}{l}{} &  & \multicolumn{1}{l}{} \\
SAPHNet \cite{suin2020spatially}                               &  & \multicolumn{1}{c}{2020} &  & 32.02                &  & 0.953                \\
\multicolumn{1}{l}{}                  &  &                          &  & \multicolumn{1}{l}{} &  & \multicolumn{1}{l}{} \\
RADNet \cite{purohit2020region}                               &  & \multicolumn{1}{c}{2020} &  & 32.15                &  & 0.953                \\
\multicolumn{1}{l}{}                  &  &                          &  & \multicolumn{1}{l}{} &  & \multicolumn{1}{l}{} \\
BANet \cite{tsai2021banet}                             &  & \multicolumn{1}{c}{2021} &  & 32.44                &  & 0.957                \\
\multicolumn{1}{l}{}                  &  &                          &  & \multicolumn{1}{l}{} &  & \multicolumn{1}{l}{} \\
MPRNet \cite{zamir2021multi}                              &  & \multicolumn{1}{c}{2021} &  & 32.66                &  & 0.959                \\
\multicolumn{1}{l}{}                  &  &                          &  & \multicolumn{1}{l}{} &  & \multicolumn{1}{l}{} \\
MIMO-UNet++ \cite{cho2021rethinking}                           &  & \multicolumn{1}{c}{2021} &  & 32.68                &  & 0.959                \\
\multicolumn{1}{l}{}                  &  &                          &  & \multicolumn{1}{l}{} &  & \multicolumn{1}{l}{} \\
HINet \cite{chen2021hinet}                                 &  & \multicolumn{1}{c}{2021} &  & 32.71                &  & 0.959                \\
\multicolumn{1}{l}{}                  &  &                          &  & \multicolumn{1}{l}{} &  & \multicolumn{1}{l}{} \\
SL-CycleGAN \textbf{(Ours)}           &  & \multicolumn{1}{c}{2021} &  & \textbf{38.087}      &  & 0.954                \\ \bottomrule
\end{tabular}%
}
 \caption{Quantitative comparison of Blind image deblurring on GoPro dataset \cite{nah2017deep}. Our proposed method SL-CycleGAN achieves the highest PSNR of 38.087 dB on blind image motion deblurring task.}
 \label{Tab1}
\end{table}

\begin{figure*}[ht]
    \centering
    \includegraphics[width=\textwidth]{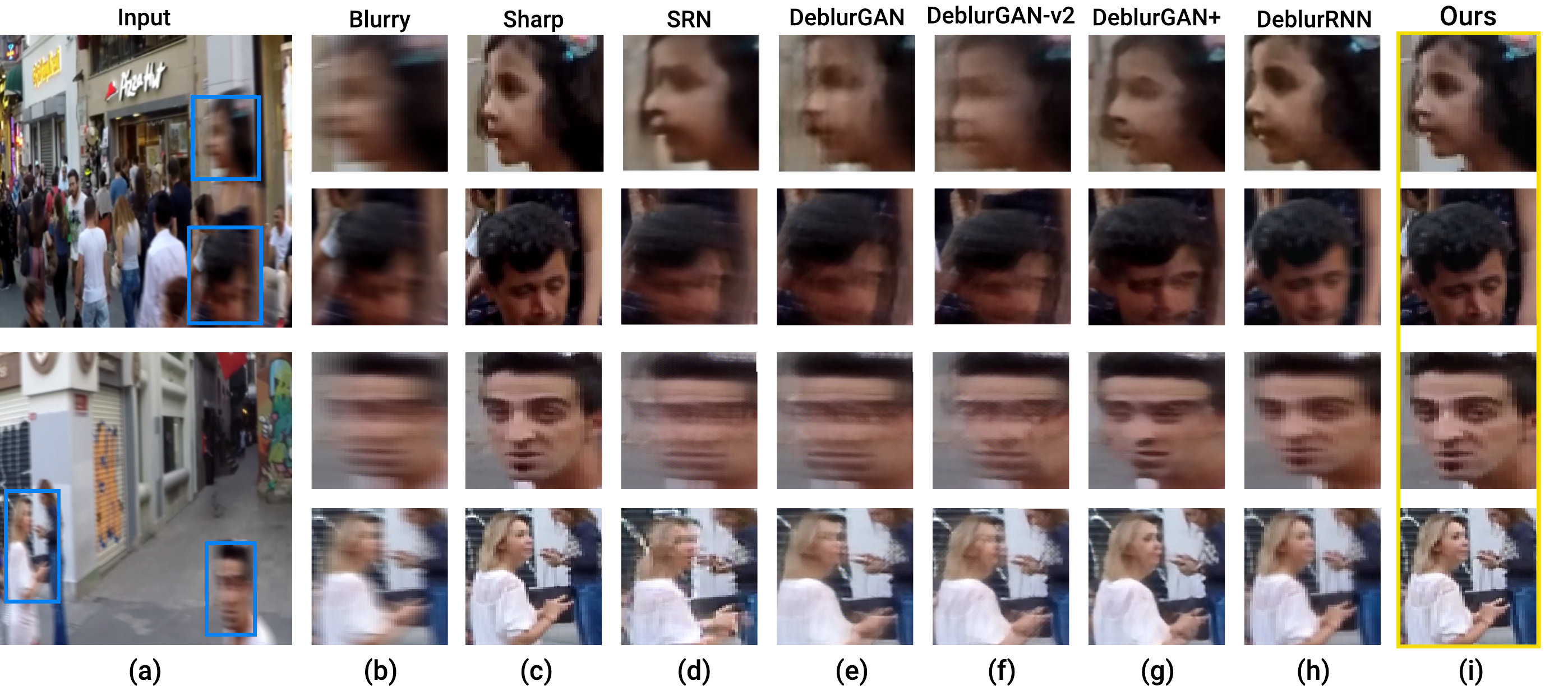}
    \caption{Deblurring results of test images from GoPro dataset. (a) Blurry inputs. (b) Magnified blurry image patches. (c) Corresponding sharp image patches. (d) Deblurring results of \cite{tao2018scale}. (e) Deblurring results of \cite{kupyn2018deblurgan}. (f) Deblurring results of \cite{kupyn2019deblurgan}. (g) Deblurring results of \cite{shao2020deblurgan+}. (h) Deblurring results of \cite{zhang2021deep}. (i) Finally, deblurring results of our proposed SL-CycleGAN.}
    \label{fig3}
\end{figure*}
\begin{figure*}[t]
    \centering
    \includegraphics[width=\textwidth]{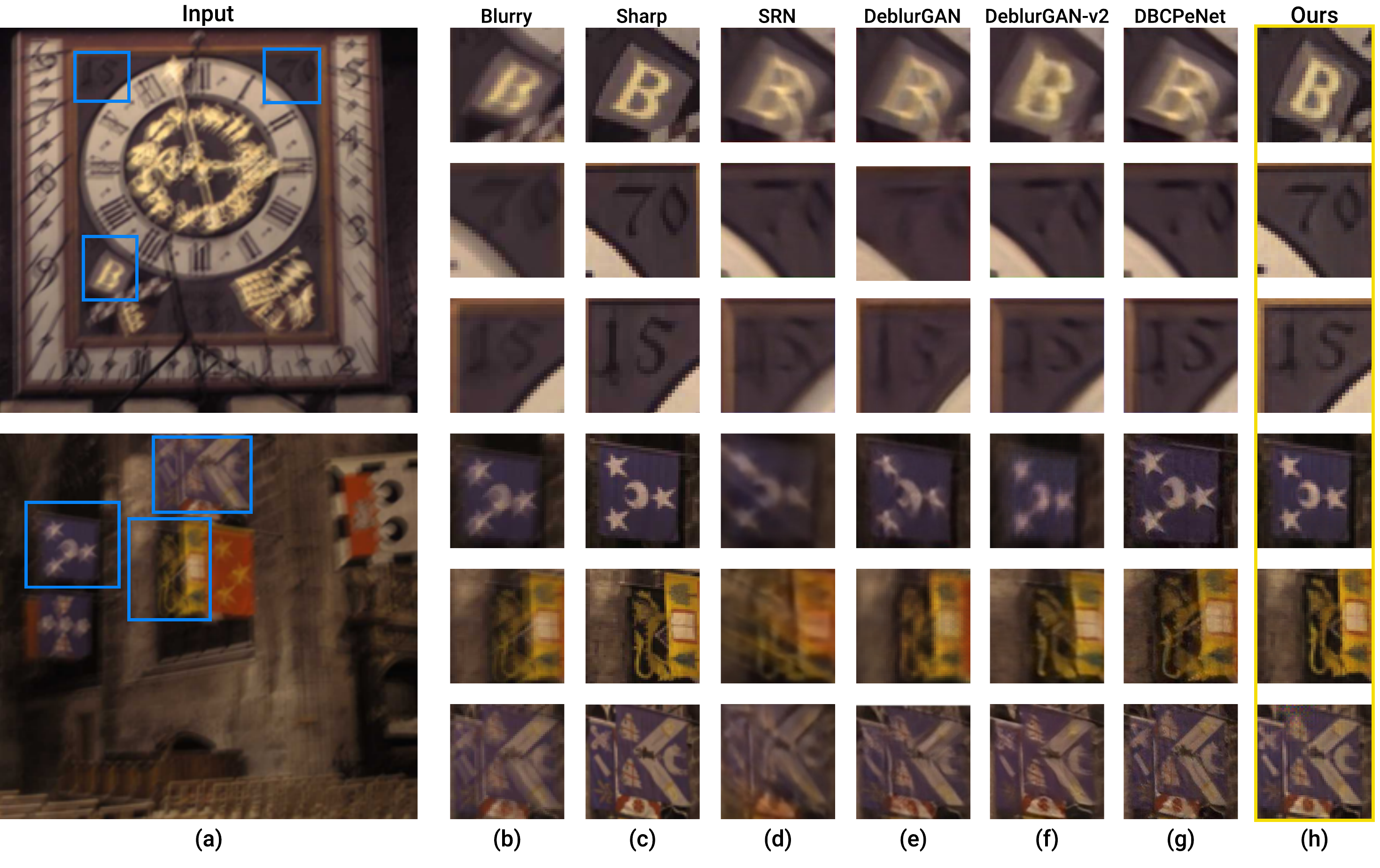}
    \caption{Visual comparison of test images from Kohler dataset \cite{kohler2012recording}. (a) Input blurry images. (b) Magnified blurry image patches. (c) Magnified real sharp image patches. (d) Deblurring results of \cite{tao2018scale}. (e) Deblurring results of \cite{kupyn2018deblurgan}. (f) Deblurring results of \cite{kupyn2019deblurgan}. (g) Deblurring results of \cite{cai2020dark}. (h) Deblurring results of our proposed method.}
    \label{fig4}
\end{figure*}
\subsection{Experiment settings}
We have used Pytorch \cite{NEURIPS2019_9015} for all our experiments on Nvidia GTX 1080ti with 11G GPU. We performed experiments on three image benchmarks, GoPro dataset \cite{nah2017deep}, Kohler dataset \cite{kohler2012recording}, and Lai dataset \cite{lai2016comparative}. We resized all the images in all three image benchmarks to 256$\times$256 for training and testing and apply data augmentation. For optimization of the generators and the discriminators, we use the Adam optimizer \cite{kingma2014adam} with ${{\beta}=0.999}$ and batch size of 1. We train our model on all these three image benchmarks for 200 training epochs each with an initial learning rate of 0.0002 for first 100 epochs and linearly decay to zero over next 100 iterations. For all the experiments we set the values of ${{\lambda}_{cyc}}=10$ and ${{\lambda }_{perc}}=100$ in \cref{eq7}. We use the gradient penalty term of \cite{gulrajani2017improved} for the discriminator networks, which is set to 10. We do not use dropout layer in our modified Sparse ResNet architecture, since \cite{ahmad2019can} in their research show that the utilization of k-winner with sparse convolutions replaces the need of dropout layers in the network architecture. The training time of our proposed network (SL-CycleGAN) on one dataset for total of 200 training epochs took 2 days to complete, which is 6 days in total for three datasets.  
\begin{table}[t]
\centering
\resizebox{7cm}{!}{%
\begin{tabular}{@{}clclc@{}}
\toprule
Method            &  & PSNR  &  & SSIM   \\ \midrule
Whyte \etal. \cite{whyte2012non}            &  & 27.02 &  & 0.809  \\
Xu \etal. \cite{xu2013unnatural}              &  & 27.40 &  & 0.810  \\
Sun \etal. \cite{sun2015learning}             &  & 25.21 &  & 0.772  \\
DeepDeblur \cite{nah2017deep}            &  & 26.48 &  & 0.807  \\
DeblurGAN \cite{kupyn2018deblurgan}            &  & 25.86 &  & 0.802  \\
DeblurGAN-v2 \cite{kupyn2019deblurgan}      &  & 26.10 &  & 0.816  \\
SRN-Deblur \cite{tao2018scale}              &  & 26.75 &  & 0.837  \\
DMPHN \cite{stacked}     &  & 24.21 &  & 0.7562 \\
Zhang \etal. \cite{zhang2018dynamic}        &  & 25.71 &  & 0.800  \\
Kim \etal. \cite{hyun2013dynamic}               &  & 24.68 &  & 0.794  \\

DBCPeNet \cite{cai2020dark}              &  & 26.79 &  & 0.839  \\
SL-CycleGAN \textbf{(ours)} &  & \textbf{30.818 } &  & \textbf{0.843}   \\ \bottomrule
\end{tabular}%
}
 \caption{Quantitative comparison on Kohler dataset \cite{kohler2012recording}. Our proposed SL-CycleGAN achieves significant improvement both in terms of PSNR and SSIM. }
 \label{Tab2}
\end{table}
\begin{table}[t]
\centering
\resizebox{7cm}{!}{%
\begin{tabular}{@{}clclc@{}}
\toprule
Method            &  & PSNR   &  & SSIM  \\ \midrule
Fergus \etal. \cite{fergus2006removing}            &  & 22.870 &  & 0.682 \\
Cho \cite{cho2009fast}              &  & 23.272 &  & 0.699 \\
Xu \etal. \cite{xu2013unnatural}               &  & 25.586 &  & 0.773 \\
Krishnan \etal. \cite{krishnan2011blind}         &  & 23.070 &  & 0.716 \\
Levin \etal. \cite{levin2009understanding}             &  & 21.855 &  & 0.651 \\
Whyte \etal. \cite{whyte2012non}             &  & 23.232 &  & 0.667 \\
Sun \etal. \cite{sun2015learning}             &  & 24.649 &  & 0.756 \\
Xu \cite{xu2010two}               &  & 25.319 &  & 0.765 \\
Zhang \etal. \cite{zhang2013multi}            &  & 22.918 &  & 0.679 \\
Chakrabarti \etal. \cite{chakrabarti2010analyzing}       &  & 25.389 &  & 0.769 \\
Nah \etal. \cite{nah2017deep}              &  & 24.224 &  & 0.713 \\
Gong \etal. \cite{gong2017motion}             &  & 23.805 &  & 0.694 \\
DeblurGAN \cite{kupyn2018deblurgan}       &  & 24.561 &  & 0.741 \\
DeblurGAN-v2 \cite{kupyn2019deblurgan}      &  & 25.634 &  & 0.754 \\
SRN-Deblur \cite{tao2018scale}      &  & 25.231 &  & 0.752 \\
SL-CycleGAN \textbf{(ours)} &  & \textbf{27.935} &  & 0.766 \\ \bottomrule
\end{tabular}%
}
\caption{Quantitative comparison on Lai dataset \cite{lai2016comparative}. Our Proposed approach shows superior performance than all the other methods in terms of PSNR.}
 \label{Tab3}
\end{table}
\subsection{Image Benchmarks}
\textbf{Evaluation on GoPro Dataset:} 
GoPro dataset was proposed by Nah \etal. \cite{nah2017deep}, which consists of 3214 images in total for deblurring task, 2103 training image pairs of blurred and sharp images while the rest of 1111 images are reserved for testing purposes. It is the most commonly used benchmark for blind image deblurring task. The quantitative evaluation on GoPro dataset is presented in \cref{Tab1}. While \cref{Tab1} presents the timeline of all the state-of-the-art deep learning-based deblurring methods starting from year 2016-2021 both in terms of PSNR and SSIM. Our proposed method SL-CycleGAN outperforms all the state-of-the-art methods on GoPro deblurring task, while achieving the record-breaking PSNR of \textbf{38.087} dB, which is 5.377 dB better than the most recent deblurring method HiNet \cite{chen2021hinet}. Similarly, the average SSIM value of our proposed network remains in the list of top five most recent deblurring methods. The qualitative results on GoPro dataset are given in \cref{fig3}. In comparison with the state-of-the-art blind deblurring methods \cite{tao2018scale, kupyn2018deblurgan, kupyn2019deblurgan, shao2020deblurgan+, zhang2021deep}, our proposed approach restores the sharp images from the blurry inputs that are similar to the real sharp images and can be clearly seen in \cref{fig3}. The resemblance between our restored and the real sharp image patches is quite high in comparison with other approaches. \par
\textbf{Evaluation on Kohler Dataset:}
Kohler \etal \cite{kohler2012recording} proposed a real-world deblurring datasat that consists of 4 latent sharp images and 48 corresponding blurry images of varying blur kernel intensities. It is the most commonly used benchmark for blind image deblurring comparison. The quantitative comparison of blind image deblurring on Kohler dataset is given in \cref{Tab2}. Our SL-CycleGAN outperforms the other state-of-the-art methods by achieving the average PSNR of 30.818 dB and SSIM of 0.843, while our closest competitor DBCPeNet \cite{cai2020dark} achieves the PSNR of 26.79 dB and SSIM of 0.839. Similarly, DeblurGAN \cite{kupyn2018deblurgan}, DeblurGAN-v2 \cite{kupyn2019deblurgan}, SRN-Deblur \cite{tao2018scale} and DMPHN \cite{stacked} show quantitatively inferior performance than our proposed approach. The visual comparison of test images from kohler dataset is presented in \cref{fig4}. We can see from \cref{fig4} that our deblurred sharp image patches retain the texture details and the sharpness similar to the original latent sharp image patches. In comparison with \cite{tao2018scale, kupyn2018deblurgan, kupyn2019deblurgan, cai2020dark}, our proposed model shows the ability to understand the distribution of non-uniform blur over different image regions even when the subject of focus lacks significant light reflection. \par
\begin{figure*}[ht]
    \centering
    \includegraphics[width=\textwidth]{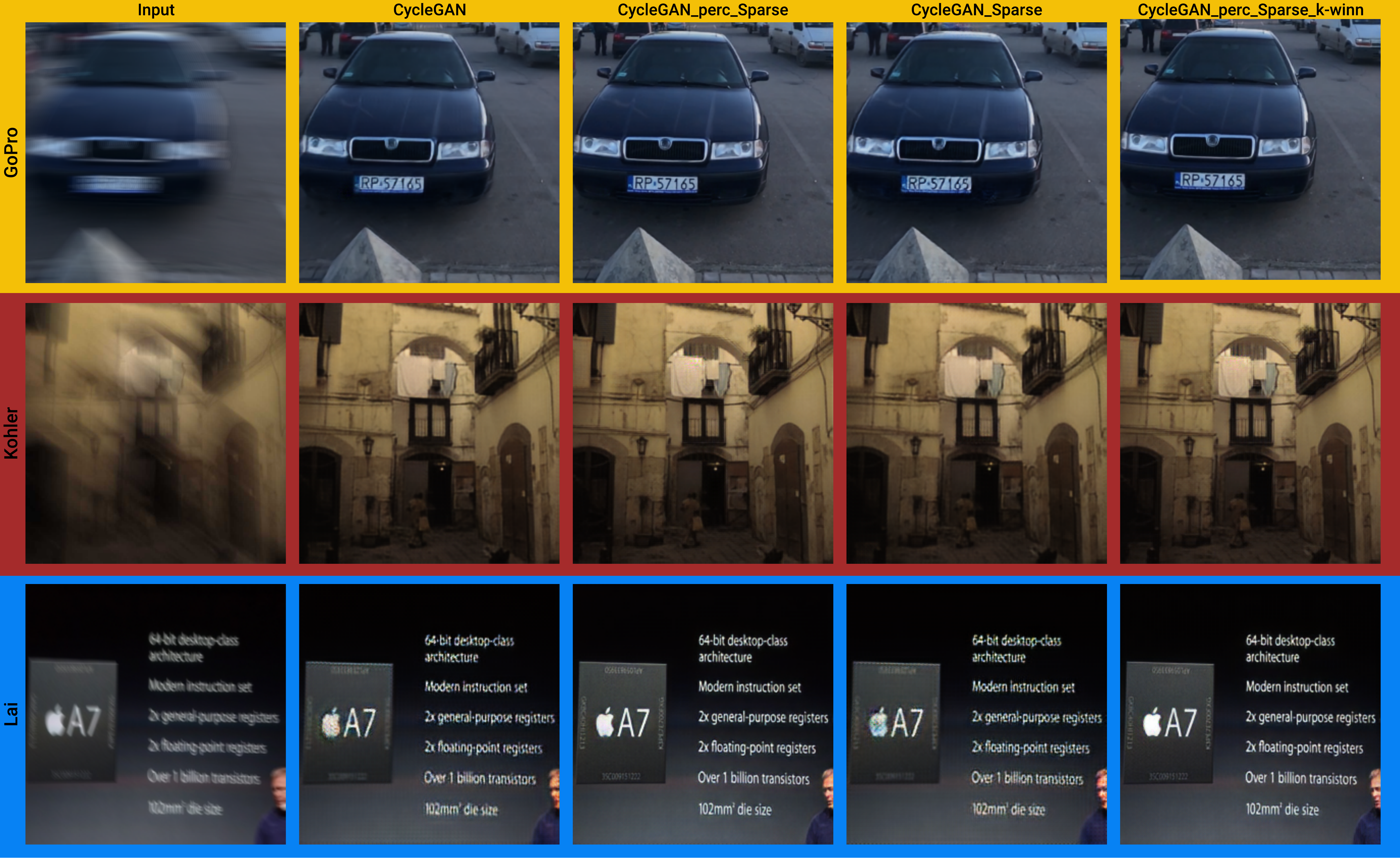}
    \caption{Visual ablation study and analysis on GoPro \cite{nah2017deep}, Kohler \cite{kohler2012recording}, and Lai \cite{lai2016comparative} datasets. First row: Images from GoPro. Second row: Images from Kohler dataset. Third row: Images from Lai dataset. Meanwhile the first column represents the blurry inputs, second column: represent deblur results of CycleGAN, third column: CycleGAN with perceptual loss and sparse convs, fourth column: CycleGAN and sparse convs, and finally CycleGAN + perceptual loss + sparse + k-winn (SL-CycleGAN). }
    \label{fig5}
\end{figure*}
\textbf{Quantitative evaluation on Lai Dataset:}
Lai \etal. \cite{lai2016comparative} proposed a benchmark for blind image deblurring task, which contains 100 real-world blurred images, they also generated synthetic dataset with 200 generated blurred images containing images of both uniform and non-uniform blur. We present the quantitative comparison on Lai dataset in \cref{Tab3}. We can observe from \cref{Tab3} that our proposed method achieves 2.301 dB improvement in PSNR than the second highest DeblurGAN-v2 \cite{kupyn2019deblurgan}. Meanwhile, the visual results based on the ablation study and analysis on Lai dataset are shown in \cref{fig5}, which we further discuss in \cref{ablation} along with the ablation analysis on GoPro and Kohler datasets. 
\begin{table}[t]
\centering
\resizebox{\columnwidth}{3cm}{%
\begin{tabular}{@{}cccc@{}}
\toprule
GoPro                                             & \multicolumn{1}{l}{PSNR(dB)} & \multicolumn{1}{l}{SSIM} & \multicolumn{1}{l}{MS-SSIM} \\ \midrule
CycleGAN                                          & 31.835                       & 0.844                    & 0.986                       \\
CycleGAN+VGG-19(perceptual)+Sparse                           & 37.852                      & 0.954                  & 0.997                       \\
CycleGAN+Sparse                                   & 33.135                       & 0.876                    & 0.990                       \\
CycleGAN+VGG-19(perceptual)+Sparse+k-winn (SL-CycleGAN)                    & \textbf{38.087}                       & 0.954                   & 0.997                      \\ \hline
Kohler                                               & \multicolumn{3}{l}{}                                                                  \\ \hline                                                               
CycleGAN                                          & 29.870                       & 0.814                    & 0.983                       \\
CycleGAN+VGG-19(perceptual)+Sparse                           & 29.985                       & 0.814                    & 0.984                       \\
CycleGAN+Sparse                                   & 30.461                       & 0.823                    & 0.985                       \\
\multicolumn{1}{l}{CycleGAN+VGG-19(perceptual)+Sparse+k-winn} (SL-CycleGAN) & \textbf{30.818}                       & \textbf{0.843 }                   & \textbf{0.987}                      \\ \hline
Lai                                               & \multicolumn{3}{l}{}                                                                  \\ \hline
CycleGAN                                          & 25.034                       & 0.662                    & 0.970                       \\
CycleGAN+VGG-19(perceptual)+Sparse                           & 27.581                       & 0.764                    & 0.983                       \\
CycleGAN+Sparse                                   & 27.564                       & 0.757                    & 0.983                       \\
\multicolumn{1}{l}{CycleGAN+VGG-19(perceptual)+Sparse+k-winn} (SL-CycleGAN) & \textbf{27.935}                       & \textbf{0.766}                    & \textbf{0.984}                      \\ \bottomrule
\end{tabular}%
}
\caption{Quantitative ablation study on GoPro \cite{nah2017deep}, Kohler \cite{kohler2012recording} and Lai \cite{lai2016comparative} datasets. }
 \label{Tab4}
\end{table}
\subsection{Ablation Study}\label{ablation}
We conduct an ablation study on the components of SL-CycleGAN and observe the impact and effectiveness of these components both qualitatively and quantitatively. We present the visual ablation study and on three image benchmarks in \cref{fig5}, while considering original CycleGAN \cite{zhu2017unpaired} as an starting point. Meanwhile, we gradually keep adding modifications to the generator networks such as replacing standard conv layers in the ResNet by sparse-convs and adding VGG-19 perceptual loss, then eliminating perceptual loss and leaving only sparse conv layers. Finally, we modify the network by integrating perceptual loss, sparse-convs and replace ReLU with k-winner in the ResNet generators architecture. We call the final version of our network SL-CycleGAN (CycleGAN + perceptual + sparse-convs + k-winn). We can see from \cref{fig5} that all our sparse versions of the network perform better visually than just CycleGAN, especially the final version SL-CycleGAN produces visually appealing results. Similarly in \cref{Tab4} of ablation study, SL-CycleGAN outperforms all the preceding versions quantitatively.\par
\textbf{Limitations:} Keeping in mind that ``Honesty is the best policy''. We observe that during the inference on GoPro dataset, some of the restored images by SL-CyleGAN show slightly dim light in comparison with the original bright sharp images. However, we consider it as an inherited issue from the original CycleGAN and cycle-consistency loss \cite{zhu2017unpaired}. 
\section{Conclusion}
This paper introduces a novel blind image deblurring network SL-CycleGAN, that for the first time, utilizes sparse representation learning with HTM k-winner for improved image deblurring and is more robust towards noise and interference. Meanwhile, achieving the best qualitative and quantitative results on popular image benchmarks.











{\small
\bibliographystyle{ieee_fullname}
\bibliography{egbib}
}

\end{document}